\documentclass[sigconf]{acmart}

\usepackage{natbib}
\usepackage[all]{nowidow}

\AtBeginDocument{%
  \providecommand\BibTeX{{%
    \normalfont B\kern-0.5em{\scshape i\kern-0.25em b}\kern-0.8em\TeX}}}

\usepackage{amsmath}
\usepackage{amssymb}
\renewcommand{\vec}[1]{\boldsymbol{#1}}
\DeclareMathOperator*{\argmin}{arg\,min}

\copyrightyear{2020}
\acmYear{2020}
\setcopyright{rightsretained}
\acmConference[FAT* '20]{Conference on Fairness, Accountability, and Transparency}{January 27--30, 2020}{Barcelona, Spain}
\acmBooktitle{Conference on Fairness, Accountability, and Transparency (FAT* '20), January 27--30, 2020, Barcelona, Spain}\acmDOI{10.1145/3351095.3375624}
\acmISBN{978-1-4503-6936-7/20/02}


\begin{document}

\title[Deploying Explainability]{Explainable Machine Learning in Deployment}


  



  
  

  
  
\author{Umang Bhatt$^{1,2,3,4}$, Alice Xiang$^{2}$, Shubham Sharma$^{5}$, Adrian Weller $^{3,4,6}$, Ankur Taly$^{7}$, Yunhan Jia$^{8}$, Joydeep Ghosh$^{5,9}$, Ruchir Puri$^{10}$, Jos\'e M. F. Moura$^{1}$, Peter Eckersley$^{2}$}
\affiliation{
  \institution{$^{1}$Carnegie Mellon University, $^{2}$Partnership on AI, $^{3}$University of Cambridge, $^{4}$Leverhulme CFI, $^{5}$University of Texas at Austin, $^{6}$The Alan Turing Institute, $^{7}$Fiddler Labs, $^{8}$Baidu, $^{9}$CognitiveScale, $^{10}$IBM Research}
}

\renewcommand{\shortauthors}{Bhatt et al.} 

\begin{abstract}
Explainable machine learning offers the potential to provide stakeholders with insights into model behavior by using various methods such as feature importance scores, counterfactual explanations, or influential training data. Yet there is little understanding of how organizations use these methods in practice. 
This study explores how organizations view and use explainability for stakeholder consumption. 
We find that, currently, the majority of deployments are not for end users affected by the model but rather for machine learning engineers, who use explainability to debug the model itself. There is thus a gap between explainability in practice and the goal of transparency, since explanations primarily serve internal stakeholders rather than external ones.
Our study synthesizes the limitations of current explainability techniques that hamper their use for end users.
To facilitate end user interaction, we develop a framework for establishing clear goals for explainability. We end by discussing concerns raised regarding explainability.
\end{abstract}


\maketitle

\section{Introduction}
Machine learning (ML) models are being increasingly embedded into many aspects of daily life, such as healthcare \cite{de2018clinically}, finance \cite{heaton2016deep}, and social media \cite{alvarado2018towards}. To build ML models worthy of human trust, researchers have proposed a variety of techniques for explaining ML models to stakeholders. Deemed ``explainability,'' this body of previous work attempts to illuminate the reasoning used by ML models. ``Explainability'' loosely refers to any technique that helps the user or developer of ML models understand why models behave the way they do. Explanations can come in many forms: from telling patients which symptoms were indicative of a particular diagnosis \cite{lundberg2018explainable} to helping factory workers analyze inefficiencies in a production pipeline \cite{dhurandhar2018improving}. 

Explainability has been touted as a way to enhance transparency of ML models  \cite{lepri2018fair}. Transparency includes a wide variety of efforts to provide stakeholders, particularly end users, with relevant information about how a model works \cite{weller2019transparency}. One form of this would be to publish an algorithm's code, though this type of transparency would not provide an intelligible explanation to most users. Another form would be to disclose properties of the training procedure and datasets used \cite{mitchell2019model}. 
Users, however, are generally not equipped to be able to understand how raw data and code translate into benefits or harms that might affect them individually. By providing an explanation for how the model made a decision, explainability techniques seek to provide transparency directly targeted to human users, often aiming to increase trustworthiness \cite{o2018linking}. The importance of explainability as a concept has been reflected in legal and ethical guidelines for data and ML \cite{selbst2018intuitive}. In cases of automated decision-making, Articles 13-15 of the European General Data Protection Regulation (GDPR) require that data subjects have access to ``meaningful information about the logic involved, as well as the significance and the envisaged consequences of such processing for the data subject'' \cite{gdpr}. In addition, technology companies have released artificial intelligence (AI) principles that include transparency as a core value, including notions of explainability, interpretability, or intelligibility \cite{ibm2019,msft2019}. 

With growing interest in ``peering under the hood'' of ML models and in providing explanations to human users, explainability has become an important subfield of ML.
Despite a burgeoning literature, there has been little work characterizing how explanations have been deployed by organizations in the real world. 

In this paper, we explore how organizations have deployed local explainability techniques so that we can observe which techniques work best in practice, report on the shortcomings of existing techniques, and recommend paths for future research. 
We focus specifically on local explainability techniques since these techniques explain individual predictions, making them typically the most relevant form of model transparency for end users.

Our study synthesizes interviews with roughly fifty people from approximately thirty organizations to understand which explainability techniques are used and how.
We report trends from two sets of interviews and provide recommendations to organizations deploying explainability. 
To the best of our knowledge, we are the first to conduct a study of how explainability techniques are used by organizations that deploy ML models in their workflows.
Our main contributions are threefold:
\begin{itemize}
    \item We interview around twenty data scientists, who are not currently using explainability tools, to understand their organization's needs for explainability.
    \item We interview around thirty different individuals on how their organizations have deployed explainability techniques, reporting case studies and takeaways for each technique.
    \item We suggest a framework for organizations to clarify their goals for deploying explainability.
\end{itemize}

The rest of this paper is organized as follows:
\begin{enumerate}
    \item We discuss the methodology of our survey in Section~\ref{sec:prelim}.
    \item We summarize our overall findings in Section~\ref{sec:summary}.
    \item We detail how local explainability techniques are used at various organizations and discuss technique-specific takeaways in Section~\ref{sec:use-local}. 
    \item We develop a framework for establishing clear goals when deploying local explainability in Section~\ref{sec:desire} and discuss concerns of explainability in Section~\ref{sec:trends}.
\end{enumerate}

\section{Methodology}
\label{sec:prelim}

In the spirit of \citet{holstein2019improving}, we study how industry practitioners look at and deploy explainable ML. Specifically, we study how particular organizations deploy explainability algorithms, including who consumes the explanation and how it is evaluated for the intended stakeholder. We conduct two sets of interviews: Group 1 consisted at how data scientists who are not currently using explainable machine learning hope to leverage various explainability tools, while Group 2, the crux of this paper, consisted at how explainable machine learning has been deployed in practice.

For Group 1, Fiddler Labs led a set of around twenty interviews to assess explainability needs across various organizations in the technology and financial services sectors. We specifically focused on teams that do not currently employ explainability tools. These semi-structured, hour-long interviews included, but were not limited to, the following questions:
\begin{itemize}
    \item What are your ML use cases?
    \item What is your current model development workflow?
    \item What are your pain points in deploying ML models?
    \item Would explainability help address those pain points?
\end{itemize}

Group 2 spanned roughly thirty people across approximately twenty different organizations, both for-profit and non-profit. Most of these organizations are members of the Partnership on AI, which is a global multistakeholder non-profit established to study and formulate best practices for AI to benefit society.
With each individual, we held a thirty-minute to two-hour semi-structured interview to understand the state of explainability in their organization, their motivation for using explanations, and the benefits and shortcomings of the methods used. Some organizations asked to stay anonymous, not to be referred to explicitly in the prose, or not to be included in the acknowledgements.

Of the people we spoke with in Group 2, around one-third represented non-profit organizations (academics, civil societies, and think tanks), while the rest worked for for-profit organizations (corporations, industrial research labs, and start-ups). Broken down by organization, around half were for-profit and half were academic / non-profit. Around one-third of the interviewees were executives at their organization, around half were research scientists or engineers, and the remainder were professors at academic institutions, who commented on the consulting they had done with industry leaders to commercialize their research. 
The questions we asked Group 2 included, but were not limited to, the following:

\begin{itemize}
    \item  Does your organization use ML model explanations?
    \item What type of explanations have you used (e.g., feature-based, sample-based, counterfactual, or natural language)?
    \item Who is the audience for the model explanation (e.g., research scientists, product managers, domain experts, or users)?
    \item In what context have you deployed the explanations (e.g., informing the development process, informing human decision-makers about the model, or informing the end user on how actions were taken based on the model's output)?
    \item How does your organization decide when and where to use model explanations?
\end{itemize}

\section{Summary of Findings}
\label{sec:summary}
Here we synthesize the results from both interview groups.
For the sake of clarity, we define various terms based on the context in which they appear in the forthcoming prose. 
\begin{itemize}
    \item \textit{Trustworthiness} refers to the extent to which stakeholders can reasonably trust a model's outputs.
    \item \textit{Transparency} refers to attempts to provide stakeholders (particularly external stakeholders) with relevant information about how the model works: this includes documentation of the training procedure, analysis of training data distribution, code releases, feature-level explanations, etc.
    \item \textit{Explainability} refers to attempts to provide insights into a model's behavior.
    \item \textit{Stakeholders} are the people who either want a model to be ``explainable,'' will consume the model explanation, or are affected by decisions made based on model output.
    \item \textit{Practice} refers to the real-world context in which the model has been deployed.
    \item \textit{Local Explainability} aims to explain the model's behavior for a specific input.
    \item \textit{Global Explainability} attempts to understand the high-level concepts and reasoning used by a model.
\end{itemize}

\subsection{Explainability Needs}
\label{sec:needs}
This subsection provides an overview of explainability needs that were uncovered with Group 1, data scientists from organizations that do not currently deploy explainability techniques. These data scientists were asked to describe their ``pain points'' in building and deploying ML models, and how they hope to use explainability.
\begin{itemize}
    \item \textbf{Model debugging}: Most data scientists struggle with debugging poor model performance. They wish to identify why the model performs poorly on certain inputs, and also to identify regions of the input space with below average performance. In addition, they seek guidance on how to engineer new features, drop redundant features, and gather more data to improve model performance.  For instance, one data scientist said: ``If I have 60 features, maybe it's equally effective if I just have 5 features.'' Dealing with feature interactions was also a concern, as the data scientist continued, ``Feature A will impact feature B, [since] feature A might negatively affect feature B---how do I attribute [importance in the presence of] correlations?'' Others mentioned explainability as a debugging solution, helping to ``narrow down where things are broken.''
    \item \textbf{Model monitoring}: Several individuals worry about drift in the feature and prediction distributions after deployment. Ideally, they would like to be alerted when there is a significant drift relative to the training distribution  \cite{amodei2016concrete,pinto2019automatic}. One organization would like explanations for how drift in feature distributions would impact model outcomes and feature contribution to the model: ``We can compute how much each feature is drifting, but we want to cross-reference [this] with which features are impacting the model a lot.'' 
    \item \textbf{Model transparency}: Organizations that deploy models to make decisions that directly affect end users seek explanations for model predictions. The explanations are meant to increase model transparency and comply with current or forthcoming regulations. In general, data scientists believe that explanations can also help communicate predictions to a broader external audience of other business teams and customers. One company stressed the need to ``show your work'' to provide reasons on underwriting decisions to customers, and another company needed explanations to respond to customer complaints.
    \item \textbf{Model audit}: In financial organizations, due to regulatory requirements, all deployed ML models must go through an internal audit. Data scientists building these models need to have them reviewed by internal risk and legal teams. One of the goals of the model audit is to conduct various kinds of tests provided by regulations like SR 11-7~\cite{sr11-7}. An effective model validation framework should include: (1) evaluation of conceptual soundness of the model, (2) ongoing monitoring, including benchmarking, and (3) outcomes analysis, including back-testing. Explainability is viewed as a tool for evaluating the soundness of the model on various data points. Financial institutions would like to conduct sensitivity analyses, checking the impact of small changes to inputs on model outputs. Unexpectedly large changes in outputs can indicate an unstable model.
\end{itemize}

\subsection{Explainability Usage}
In Table~\ref{table:use}, we aggregate some of the explainability use cases that we received from different organizations in Group 2. For each use case, we define the domain of use (i.e., the industry in which the model is deployed), the purpose of the model, the explainability technique used, the stakeholder consuming the explanation, and how the explanation is evaluated. 
Evaluation criteria denote how the organization compares the success of various explanation functions for the chosen technique (e.g., after selecting feature importance as the technique, an organization can compare LIME \cite{ribeiro2016should} and SHAP \cite{shap} explanations via the faithfulness criterion \cite{yeh2019sensitive}).

In our study, feature importance was the most common explainability technique, and Shapley values were the most common type of feature importance explanation.  The most common stakeholders were ML engineers (or research scientists), followed by domain experts (e.g., loan officers and content moderators). Section~\ref{sec:use-local} provides definitions for each technique and further details on how these techniques were used at Group 2 organizations.

\begin{table*}
\begin{center}
\begin{small}
\begin{sc}
\begin{tabular}{ccccc}
\toprule
Domain & Model Purpose & Explainability Technique & Stakeholders & Evaluation Criteria \\
\midrule
Finance & Loan Repayment & Feature Importance & Loan Officers & Completeness \cite{shap} \\
Insurance & Risk Assessment & Feature Importance & Risk Analysts & Completeness \cite{shap} \\
Content Moderation & Malicious Reviews  & Feature Importance & Content Moderators & Completeness \cite{shap} \\
Finance & Cash Distribution & Feature Importance & ML Engineers & Sensitivity \cite{yeh2019sensitive} \\
Facial Recognition & Smile Detection & Feature Importance & ML Engineers & Faithfulness \cite{ancona2018towards}\\
Content Moderation & Sentiment Analysis & Feature Importance & QA ML Engineers & $\ell_2$ norm \\
Healthcare & Medicare access & Counterfactual Explanations & ML Engineers & normalized $\ell_1$ norm \\

Content Moderation & Object Detection & Adversarial Perturbation & QA ML Engineers & $\ell_2$ norm \\
\bottomrule
\end{tabular}
\end{sc}
\end{small}
\end{center}
\caption{Summary of select deployed local explainability use cases}
\label{table:use}
\end{table*}

\subsection{Stakeholders}
Most organizations in Group 2 deploy explainability atop their existing ML workflow for one of the following stakeholders:

\begin{enumerate}
    \item \textbf{Executives}: These individuals deem explainability necessary to achieve an organization's AI principles. One research scientist felt that ``explainability was strongly advised and marketed by higher-ups,'' though sometimes explainability simply became a checkbox.  
    \item \textbf{ML Engineers}: These individuals (including data scientists and researchers) train ML models at their organization and use explainability techniques to understand how the trained model works: do the most important features, most similar samples, and nearest training point(s) in the opposite class make sense? Using explainability to debug what the model has learned, this group of individuals were the most common explanation consumers in our study.
    \item \textbf{End Users}: This is the most intuitive consumer of an explanation. The end user is the person consuming the output of a model or making a decision based on model output. Explainability shows the end user why the model behaved the way it did, which is important for showing that the model is trustworthy and also providing greater transparency.
    \item \textbf{Other Stakeholders}: There are many other possible stakeholders for explainability. One such group is regulators, who may mandate that algorithmic decision-making systems provide explanations to affected populations or the regulators themselves. It is important that this group understands how explanations are deployed based on existing research, what techniques are feasible, and how the techniques can align with the desired explanation from a model. Another group is domain experts, who are often tasked with auditing the model's behavior and ensuring it aligns with expert intuition. For many organizations, minimizing the divergence between an expert's intuition and the model's explanation is key to successfully implementing explainability.
\end{enumerate}

Overwhelmingly, we found that local explainability techniques are mostly consumed by ML engineers and data scientists to audit models before deployment rather than to provide explanations to end users. Our interviews reveal factors that prevent organizations from showing explanations to end users or those affected by decisions made from ML model outputs.

\subsection{Key Takeaways}
\label{sec:key}
This subsection summarizes some key takeaways from Group 2 that shed light on the reasons for the limited deployment of explainability techniques and their use primarily as sanity checks for ML engineers. 
Organizations generally still consider the judgments of domain experts to be the implicit ground truth for explanations. Since explanations produced by current techniques often deviate from the understanding of domain experts, some organizations still use human experts to evaluate the explanation before it is presented to users. 
Part of this deviation stems from the potential for ML explanations to reflect spurious correlations, which result from models detecting patterns in the data that lack causal underpinnings. As a result, organizations find explainability techniques useful for helping their ML engineers identify and reconcile inconsistencies between the model's explanations and their intuition or that of domain experts, rather than for directly providing explanations to end users.

In addition, there are technical limitations that make it difficult for organizations to show end users explanations in real-time. The non-convexity of certain models make certain explanations (e.g., providing the most influential datapoints) hard to compute quickly.
Moreover, finding plausible counterfactual datapoints (that are feasible in the real world and on the input data manifold) is nontrivial, and many existing techniques currently make crude approximations or return the closest datapoint of the other class in the training set. Moreover, providing certain explanations can raise privacy concerns due to the risk of model inversion.

More broadly, organizations lack frameworks for deciding why they want an explanation, and current research fails to capture the objective of an explanation. For example, large gradients, representing the direction of maximal variation with respect to the output manifold, do not necessarily ``explain'' anything to stakeholders. At best, gradient-based explanations provide an interpretation of how the model behaves upon an infinitesimal perturbation (not necessarily a feasible one \cite{hooker2019please}), but does not ``explain'' if the model captures the underlying causal mechanism from the data. 

\section{Deploying Local Explainability}
\label{sec:use-local}
In this section, we dive into how local explainability techniques are used at various organizations (Group 2). After reviewing technical notation, we define local explainability techniques, discuss organizations' use cases, and then report takeaways for each technique.

\subsection{Preliminaries}
A black box model $\vec{f}$ maps an input $\vec{x} \in \mathcal{X} \subseteq \mathbb{R}^d$ to an output $\vec{f}(\vec{x}) \in \mathcal{Y}$, $\vec{f}: \mathbb{R}^d \mapsto \mathcal{Y}$.
When we assume $\vec{f}$ has a parametric form, we write $\vec{f}_{\theta}$. $\mathcal{L}(\vec{f}(\vec{x}), y)$ denotes the loss function used to train $\vec{f}$ on a dataset $\mathcal{D}$ of input-output pairs $(\vec{x}^{(i)}, y^{(i)})$.

Each organization we spoke with has deployed an ML model $\vec{f}$. They hope to explain a data point $\vec{x}$ using an explanation function $\vec{g}$. 
Local explainability refers to an explanation for why $\vec{f}$ predicted $\vec{f}(\vec{x})$ for a fixed point $\vec{x}$. The local explanation methods we discuss come in one of the following forms: Which feature $x_i$ of $\vec{x}$ was most important for prediction with $\vec{f}$? Which training datapoint $\vec{z} \in \mathcal{D}$ was most important to $\vec{f}(\vec{x})$? What is the minimal change to the input $\vec{x}$ required to change the output $\vec{f}(\vec{x})$?

In this paper, we deliberately decide to focus on the more popularly deployed local explainability techniques instead of global explainability techniques.
Global explainability refers to techniques that attempt to explain the model as a whole. These techniques attempt to characterize the concepts learned by the model \cite{kim2017interpretability}, simpler models learned from the representation of complex models \cite{dhurandhar2018improving}, prototypical samples from a particular model output \cite{kim2016MMD}, or the topology of the data itself \cite{dumouchel2002data}. None of our interviewees reported deploying global explainability techniques, though some studied these techniques in research settings.

\subsection{Feature Importance}
Feature importance was by far the most popular technique we found across our study. It is used across many different domains (finance, healthcare, facial recognition, and content moderation). Also known as feature-level interpretations, feature attributions, or saliency maps, this method is by far the most widely used and most well-studied explainability technique \cite{baehrens2010explain,gilpin2018explaining}.

\subsubsection{Formulation}
Feature importance methods define an explanation function $\vec{g}: \vec{f} \times \mathbb{R}^d \mapsto \mathbb{R}^d$ that takes in a model $\vec{f}$ and a point of interest $\vec{x}$ and returns importance scores $\vec{g}(\vec{f}, \vec{x}) \in \mathbb{R}^d$ for all features.
is the importance of (or attribution for) feature $x_i$ of $\vec{x}$. 

These explanation functions roughly fall into two categories: perturbation-based techniques \cite{vstrumbelj2014explaining,ribeiro2016should,shap,chen2018shapley, meaningful_pert,dasp} and gradient-based techniques \cite{smilkov2017smoothgrad,shrikumar2017learning,sundararajan2017axiomatic,ancona2018towards,dtd,second}. Note that gradient-based techniques can be seen as a special case of a perturbation-based technique with an infinitesimal perturbation size. Heatmaps are also a type of feature-level explanation that denote the importance of a region or collection of features \cite{meaningful_pert,adel2018excitation}. A prominent class of perturbation based methods is based on Shapley values from cooperative game theory \cite{shapley52}. Shapley values are a way to distribute the gains from a cooperative game to its players. In applying the method to explaining a model prediction, a cooperative game is defined between the features with the model prediction as the gain. The highlight of Shapley values is that they enjoy axiomatic uniqueness guarantees. 
Unfortunately, calculating the exact Shapley value is exponential in $d$, input dimensionality; however, the literature has proposed approximate methods using weighted linear regression~\cite{shap}, Monte Carlo approximation~\cite{vstrumbelj2014explaining}, centroid aggregation~\cite{bhatt2019towards},
and graph-structured factorization~\cite{chen2018shapley}. When we refer to Shapley-related methods hereafter, we mean such approximate methods.

\subsubsection{Shapley Values in Practice}
Organization A works with financial institutions and helps explain models for credit risk analysis. To integrate into the existing ML workflow of these institutions, Organization A proceeds as follows. They let data scientists train a model to the desired accuracy. Note that Organization A focuses mostly on models trained on tabular data, though they are beginning to venture into unstructured data (i.e., language and images). During model validation, risk analysts conduct stress tests before deploying the model to loan officers and other decision-makers. After decision-makers vet the model outputs as a sanity check and decide whether or not to override the model output, Organization A generates Shapley value explanations.

Before launching the model, risk analysts are asked to review the Shapley value explanations to ensure that the model exhibits expected behavior (i.e., the model uses the same features that a human would for the same task). Notably, the customer support team at these institutions can also use these explanations to provide individuals information about what went into the decision-making process for their loan approval or cash distribution decision. They are shown the percentage contribution to the model output (the positive $\ell_1$ norm of the Shapley value explanation along with the sign of contribution). This means that the explanation would be along the lines of, ``55\% of the decision was decided by your age, which positively correlated with the predicted outcome.''

When comparing Shapley value explanations to other popular feature importance techniques, Organization A found that in practice LIME explanations \cite{ribeiro2016should} give unexpected explanations that do not align with human intuition. Recent work \cite{badLIME} shows that the fragility of LIME explanations can be traced to the sampling variance when explaining a singular data point and to the explanation sensitivity to sample size and sampling proximity. 
Though decision-makers have access to the feature-importance explanations, end users are still not shown these explanations as reasoning for model output. Organization A aspires to eventually provide this ``explanation'' to end users.

For gradient-based language models, Organization A uses Integrated Gradients (related to  Shapley Values  by \citet{sundararajan2017axiomatic}) to flag malicious reviews and moderate content at the aforementioned institutions. This information can be highlighted to ensure the trustworthiness and transparency of the model to the decision maker (the hired content moderator here), since they can now see which word was most important to flag the content.

Going forward, Organization A intends to use a global variant of the Shapley value explanations by exposing how Shapley value explanations work on average for datapoints of a particular predicted class (e.g., on average someone who was denied a loan had their age matter most for the prediction). This global explanation would help risk analysts get a birds-eye view of how a model behaves and whether it aligns with their expectations.

\subsubsection{Heatmaps in Transportation}
Organization B looks to detect facial expressions from video feeds of users driving. They hope to use explainability to identify the actions a user is performing while the user drives. Organization B has tried feature visualization and activation visualization techniques that get attributions by backpropagating gradients to regions of interest \cite{zhang2018top,adel2018excitation}. Specifically, they use these probabilistic Winner-Take-All techniques (variants of existing gradient-based feature importance techniques \cite{shrikumar2017learning,sundararajan2017axiomatic}) to localize the region of importance in the input space for a particular classification task. For example, when detecting a smile, they expect the mouth of the driver to be important.

Though none of these desired techniques have been deployed for the end user (the driver in this case), ML engineers at Organization B found these techniques useful for qualitative review. 
On tiny datasets, engineers can figure out which scenarios have false positives (videos falsely detected to contain smiles) and why. They can also identify if true positives are paying attention to the right place or if there is a problem with spurious artifacts. 

However, while trying to understand why the model erred by analyzing similarities in false positives, they have struggled to scale this local technique across heatmaps in aggregate across multiple videos. They are able to qualitatively evaluate a sequence of heatmaps for one video, but doing so across 100M frames simultaneously is far more difficult. Paraphrasing the VP of AI at Organization B, aggregating saliency maps across videos is moot and contains little information. Note that an individual heatmap is an example of a local explainability technique, but an aggregate heatmap for 100M frames would be a global technique.  Unlike aggregating Shapley values for tabular data as done at Organization A, taking an expectation over heatmaps (in the statistical sense) does not work, since aggregating pixel attributions is meaningless. One option Organization B discussed would be to clustering low dimensional representations of the heatmaps and then tagging each cluster based on what the model is focusing on; unfortunately, humans would still have to manually label the clusters of important regions.

\subsubsection{Spurious Correlations}
Related to model monitoring for feature drift detection discussed in Section~\ref{sec:needs}, Organization B has encountered issues with spurious correlations in their smile detection  models.
Their Vice President of AI noted that ``[ML engineers] must know to what extent you want ML to leverage highly correlated data to make classifications.'' Explainability can help identify models that focus on that correlation and can find ways to have models ignore it.
For example, there may be a side effect of a correlated facial expression or co-occurrence: cheek raising, for example, co-occurs with smiling. In a cheek-raise detector trained on the same dataset as a smile detector but with different labels, the model still focused on the mouth instead of the cheeks. Both models were fixated on a prevalent co-occurrence. Attending to the mouth was undesirable in the cheek-raise detector but allowed in the smile detector.

One way Organization B combats this is by using simpler models on top of complex feature engineering. For example, they use black box deep learning models for building good descriptors that are robust across camera viewpoints and will detect different features that subject matter experts deem important for drowsiness. There is one model per important descriptor (i.e., one model for eyes closed, one for yawns, etc.). Then, they fit a simple model on the extracted descriptors such that the important descriptors are obvious for the final prediction of drowsiness. Ideally, if Organization B had guarantees about the disentanglement of data generating factors \cite{adel2018discovering}, they would be able to understand which factors (descriptors) play a role in downstream classification.


\subsubsection{Feature Importance - Takeaways}
\begin{enumerate}
    \item Shapley values are rigorously motivated, and approximate methods are  simple to deploy for decision makers to sanity check the models they have built.
    \item Feature importance is not shown to end users, but is used by machine learning engineers as a sanity check. Looping other stakeholders (who make decisions based on model outputs) into the model development process is essential to understanding what type of explanations the model delivers.
    \item Heatmaps (and feature importance scores, in general) are hard to aggregate, which makes it hard to do false positive detection at scale.
    \item Spurious correlations can be detected with simple gradient-based techniques.
\end{enumerate}

\subsection{Counterfactual Explanations}
Counterfactual explanations 
are techniques that explain individual predictions by providing a means for recourse. Contrastive explanations that highlight contextually relevant information to the model output are most similar to human explanations\cite{mittelstadt2019explaining}; however, in their current form, finding the relevant set of plausible counterfactual point is no clear. Moreover, while some existing open source implementations for counterfactual explanations exist \cite{ustun2019actionable, wexler2019if}, they either work for specific model-types or are not black-box in nature. In this section, we discuss the formulation for counterfactual explanations
and describe one solution for each deployed technique. 

\subsubsection{Formulation}
Counterfactual explanations are points close to the input for which the decision of the classifier changes. For example, for a person who was rejected for a loan by a ML model, a counterfactual explanation would possibly suggest: "Had your income been greater by \$5000, the loan would have been granted." 

Given an input $\vec{x}$, a classifier $\vec{f}$, and a distance metric $d$, we find a counterfactual explanation $\vec{c}$  by solving the optimization problem:

\begin{equation}
 \begin{aligned}
\min_{\vec{c}} d(\vec{x},\vec{c})\\
\textrm{s.t.} \vec{f}(\vec{x})\neq 
\vec{f}(\vec{c})    \\
\end{aligned}
\label{eq:opt}
\end{equation}

The method can be tailored to allow only certain relevant features to be changed. Note that the term counterfactual has a different meaning in the causality literature \cite{rubin, pearl2000causality}. Counterfactual explanations for ML were introduced by \citet{wachter2017counterfactual}. \citet{sharma2019certifai} provide details on existing techniques. 


\subsubsection{Counterfactual Explanations in Healthcare}
Organization C uses a faster version of the formulation in \citet{sharma2019certifai} to find counterfactual explanations for projects in healthcare. When people apply for Medicare, Organization C hopes to flag if a user's application has errors and to provide explanations on how to correct the errors. Moreover, ML engineers can use the robustness score to compare different models trained using this data: this robustness score is effectively a suitably normalized and averaged distance between the counterfactual and original point in Euclidean space. The original formulation makes use of a slower genetic algorithm, so they optimized the counterfactual explanation generation process. They are currently developing a first-of-its-kind application that can directly take in any black-box model and data and return a robustness score, fairness measure, and counterfactual explanations, all from a single underlying algorithm.

The use of this approach has several advantages: it can be applied to black-box models, works for any input data type, and generates multiple explanations in a single run of the algorithm. 
However, there are some shortcomings that Organization C is addressing. One challenge of counterfactual models is that the counterfactual might not be feasible. Organization C addresses this by using the training data to guide the counterfactual generation process and by providing a user interface that allows domain experts to specify constraints. The flexibility of the counterfactual approach comes with a drawback common among explanations for black-box models: there is no guarantee of the optimality of the explanation since black-box techniques cannot guarantee optimality.  

Through the creation of a deployed solution for this method, the organization realized that clients would ideally want an explainability score, along with a measure of fairness and robustness; as such, they have developed an explainability score that can be used to compare the explainability of different models.




\subsubsection{Counterfactual Explanations - Takeaways}
\begin{enumerate}
    \item Organizations are interested in counterfactual explanation solutions since the underlying method is flexible and such explanations are easy for end users to understand.
    \item It is not clear exactly what should be optimized for when generating a counterfactual or how to do it efficiently. Still, approximate solutions may suffice in practical applications.
\end{enumerate}

\subsection{Adversarial Training}
In order to ensure the model being deployed is robust to adversaries and behaves as intended, many organizations we interviewed use adversarial training to improve performance. It has recently been shown that in fact, this also can lead to more 
human interpretable features \cite{ilyas2019adversarial}.

\subsubsection{Formulation}
Other works have also explored the intersection between adversarial robustness and model interpretations \cite{yeh2019sensitive,adv2int,second,ghorbani2017interpretation,dombrowski2019explanations}. 
The claim of one of these works is that the closest adversarial example should perturb `fragile' features, enabling the model to fit to robust features (indicative of a particular class) \cite{ilyas2019adversarial}. 
The setup of feature importance in the adversarial training setting from \citet{second} is as follows: $$\begin{aligned} \vec{g}(\vec{f}, \vec{x}) =\max _{\tilde{\vec{x}}} \; \; & \mathcal{L}\left(f_{\theta^{*}}(\tilde{\vec{x}}), y\right) \\ &\|\tilde{\vec{x}}-\vec{x}\|_{0} \leq k \\ &\|\tilde{\vec{x}}-\vec{x}\|_{2} \leq \rho \end{aligned}$$

We let $|\tilde{\vec{x}} - \vec{x}|$  be the top-$k$ feature importance scores of the input, $\vec{x}$. This is similar to the adversarial example setup which is usually written in the same manner as the above (without the $\ell_0$ norm to limit the number of features that changed). It is also interesting to note that the formulation to find counterfactual explanations above matches the formulation for finding adversarial examples. \citet{sharma2019certifai} use this connection to generate adversarial examples and define a black-box model robustness score.

\subsubsection{Image Content Moderation}
Organization D moderates user-generated content (UGC) on several public platforms. Specifically, the R\&D team at Organization D developed several models to detect adult and violent content from users' uploaded images. Their quality assurance (QA) team measures model robustness to improve content detection accuracy under the threat of adversarial examples.
The robustness of a content moderation model is measured by the minimum perturbation required for an image to evade detection. Given a gradient-based image classification model $\vec{f}: \mathbb{R}^{d} \rightarrow\{1, \ldots, K\}$, and we assume $\vec{f}(\vec{x})=\operatorname{argmax}_{i}\left(Z(\vec{x})_{i}\right)$ where $Z(\vec{x}) \in \mathbb{R}^{K}$ is the final (logit) layer output, and $Z(\vec{x})_{i}$ is the prediction score for the $i$-th class. The objective can be formulated as the following optimization problem to find the minimum perturbation:
\begin{equation}
\underset{\vec{x}}{\operatorname{argmin}}\left\{d\left(\vec{x}, \vec{x}_{0}\right)+c \mathcal{L}(\vec{f}(\vec{x}), y)\right\}
\end{equation}
$d(\cdot, \cdot)$ is some distance measure that Organization D chooses to be the $\ell_2$ distance in Euclidean space; $\mathcal{L}(\cdot)$ is the cross-entropy loss function and $c$ is a balancing factor.

As is common in the adversarial literature, Organization D applies Projected Gradient Descent (PGD) to search for the minimum perturbation from the set of allowable perturbations $\mathcal{S} \subseteq \mathbb{R}^d$ \cite{madry2017towards}. The search process can be formulated as $$\vec{x}^{t+1}=\Pi_{\vec{x}+\mathcal{S}}\left(\vec{x}^{t}+\alpha \operatorname{sgn}\left(\nabla_{\vec{x}} \mathcal{L}\left(f_{\theta^{*}}(\vec{x}), y\right)\right)\right)$$ until $x^t$ is misclassified by the detection model. ML engineers on the QA team are shown a $\ell_2$-norm perturbation distance averaged over $n$ test images randomly sampled from the test dataset. The larger the average perturbation, the more robust the model is, as it takes greater effort for an attacker to evade detection. The average perturbation required is widely used as a metric when comparing different candidate models and different versions of a given model.

Organization D finds that more robust models have more convincing gradient-based explanations, i.e., the gradient of the output with respect to the input shows that the model is focusing on relevant portions of the images, confirming recent research \cite{tsipras2018robustness,adv2int,ilyas2019adversarial}.

\subsubsection{Text Content Moderation}
Organization E uses text content moderation algorithms on its UGC platforms, such as forums. Its QA team is responsible for the reliability and robustness of a sentiment analysis model, which labels posts as positive or negative, trained on UGC. The QA team seeks to find the minimum perturbation required to change the classification of a post. In particular, they want to know how to take misclassified posts (e.g., negative ones classified as positive) and change them to the correct class. 

Given a sentiment analysis model $\vec{f} : \mathcal{X} \rightarrow \mathcal{Y}$, which maps from feature space $\mathcal{X}$ to a set of class $\mathcal{Y}$, an adversary aims to generate an adversarial post $\vec{x}_{adv}$ from the original post $\vec{x} \in \mathcal{X}$ whose ground truth label is $\vec{f}(\vec{x}) = \boldsymbol{y} \in \mathcal{Y}$ so that $\vec{f}\left(\vec{x}_{\boldsymbol{a} \boldsymbol{d} \boldsymbol{v}}\right)\neq y$. The QA team tries to minimize $d(\vec{x}, \vec{x}_{adv})$ for a domain-specific distance function. Organization E uses the $\ell_2$ distance in the embedding space, but it is equally valid to use the editing distance \cite{niu2018word}. Note that perturbation technique changes accordingly. 

In practice, to find the minimum distance in embedding space, Organization E chooses to iteratively modify the words in the original post, starting from the words with the highest importance. Here importance is defined as the gradient of the model output with respect to a particular word. ML engineers compute the Jacobian matrix of the given posts $\vec{x}=\left(x_{1}, x_{2}, \cdots, x_{N}\right)$ where $x_i$ is the $i$-th word. The Jacobian matrix is as follows:
\begin{equation}
    J_{\vec{f}}(\vec{x})=\frac{\partial \vec{f}(\vec{x})}{\partial \vec{x}}=\left[\frac{\partial \vec{f}_{j}(\vec{x})}{\partial x_{i}}\right]_{i \in 1 \ldots N, j \in 1 \ldots K}
\end{equation}
where $K$ represents the number of classes (in this case $K=2$), and $\vec{f}_{j}(\cdot)$ represents the confidence value of the $j$th class. The importance of word $x_i$ is defined as 
\begin{equation}
    C_{x_{i}}=J_{\vec{f},i, y}=\frac{\partial \vec{f}_{y}(\vec{x})}{\partial x_{i}}
\end{equation}
i.e., the partial derivative of the confidence value based on the predicted class $y$ regarding to the input word $x_i$. This procedure ranks the words by their impact on the sentiment analysis results. The QA team then applies a set of transformations/perturbations to the most important words to find the minimum number of important words that must be perturbed in order to flip an sentiment analysis API result. 

\subsubsection{Adversarial Training - Takeaways}
\begin{enumerate}
    \item There is a relation between model robustness and explainability. Model robustness improves the quality of feature importances (specifically saliency maps), confirming recent research findings \cite{adv2int}.
    \item Feature importance helps find minimal adversarial perturbations for language models in practice.
\end{enumerate}

\subsection{Influential Samples}
This technique asks the question: Which data point in the training dataset $\vec{x} \in \mathcal{D}_x$ is most influential to the model's output $\vec{f}(\vec{x}_{\text{test}})$ for a test point $\vec{x}_{\text{test}}$? Statisticians have used measures like Cook's distance \cite{cook1977detection} which measure the effect of deleting a data point on the model output. However, such measures require an exhaustive search and hence do not scale well for larger datasets.

\subsubsection{Formulation}
For over half of the organizations, influence functions has been the tool of choice for explaining which training points are influential to the model's output for a point $\vec{x}$ \cite{koh2017understanding} (though only one organization actually deployed the technique). We let $\mathcal{L}(\vec{f}_{\theta}, \vec{x})$ be the model's loss for point $\vec{x}$. 
The empirical risk minimizer is given by $\vec{\hat f}_{\theta} = \argmin_{\theta \in \Theta} \frac{1}{N}\sum_{i=1}^{N}\mathcal{L}(\vec{f}_{\theta}, y_{\vec{x}^{(i)}})$. Note that $y_{\vec{x}} = \vec{\hat f}_{\theta}(\vec{x})$ is the predicted output at $\vec{x}$ with the trained risk minimizer.
\citet{koh2017understanding} define the most influential data point $\vec{z}$ to a fixed point $\vec{x}$ as that which maximizes the following:
$$
\mathcal{I}_{\mathrm{up}, \operatorname{loss}}\left(\vec{z}, \vec{x}\right) = -\nabla_{\theta} \mathcal{L}\left(\vec{\hat f}_{\theta}(\vec{x}), y_{\vec{x}}\right)^{\top} H_{\vec{\hat f}_{\theta}}^{-1} \nabla_{\theta} \mathcal{L}\left(\vec{\hat f}_{\theta}(\vec{z}), y_{\vec{x}}\right)
$$
This quantity measures the effect of upweighting on datapoint ($\vec{z}$) on the loss at $\vec{x}$. The goal of sample importance is to uncover which training examples, when perturbed, would have the largest effect (positive or negative) on the loss of a test point. 

\subsubsection{Influence Functions in Insurance}
Organization F uses influence functions to explain risk models in the insurance industry. They hope to identify which customers might see an increase in their premiums based on their driving history in the past. The organization hopes to divulge to the end user how the premiums for drivers similar to them are priced. In other words, they hope to identify the influential training data points \cite{koh2017understanding} to understand which past drivers had the greatest influence on the prediction for the observed driver. Unfortunately, Organization F has struggled to provide this information to end users since the Hessian computation has made doing so impractical since the latency is high. 

More pressingly, even when Organization F lets the influence function procedure run, they find that many influential data points are simply outliers that are important for all drivers since those anomalous drivers are far out of distribution. As a result, instead of identifying which drivers are most similar to a given driver, the influential sample explanation identifies drivers that are very different from any driver (i.e., outliers). While this is could in theory be useful for outlier detection, it prevents the explanations from being used in practice. 

\subsubsection{Influential Samples - Takeaways}
\begin{enumerate}
    \item Influence functions can be intractable for large datasets; as such, a significant effort is needed to improve these methods to make them easy to deploy in practice.
    \item Influence functions can be sensitive to outliers in the data, such that they might be more useful for outlier detection than for providing end users explanations.
\end{enumerate}

\section{Recommendations}
This section provides recommendations for organizations based on the key takeaways in Section~\ref{sec:key} and the technique-specific takeaways in Section~\ref{sec:use-local}. In order to address the challenges organizations face when striving to provide explanations to end users, we recommend a framework for establishing clear desiderata in explainability and then include concerns associated with explainability.

\subsection{Establish Clear Desiderata}
\label{sec:desire}
Most organizations we spoke to solely deploy explainability techniques for internal engineers and scientists, as a debugging mechanism or as a sanity check. At the same time, these organizations  affirmed the importance of understanding the stakeholder, and hope to be able to explain a model prediction to the end user.
Once the target population of the explanation is understood, organizations can devise and deploy explainability techniques accordingly.  
We propose the following three steps for establishing clear desiderata and improving decision making around explainability. These include:  clearly identifying the target population, understanding their needs, and clarifying the intention of the explanation.

\begin{enumerate}
    \item \textbf{Identify stakeholders.} Who are your desired explanation consumers? Typically this will be those affected by or shown model outputs. \citet{preece2018stakeholders} describe how stakeholders have different needs for explainability. Distinctions between these groups can help design better explanation techniques.
    \item \textbf{Engage with each stakeholder.} Ask the stakeholder some variant of ``What would you need the model to explain to you in order to understand, trust, or contest the model prediction?'' and ``What type of explanation do you want from your model?'' \citet{doshi2017towards} highlight how the task being modeled dictates what type of explanation the human will need from the model.
    \item \textbf{Understand the purpose of the explanation.} Once the context and utility of the explanation are stated, understand what the stakeholder wants to do with the explanation \cite{gilpin2018explaining}.
    \begin{itemize}
        \item \textit{Static Consumption}: Will the explanation be used as a one-off sanity check for some stakeholders or shown to other stakeholders as reasoning for a particular prediction
        \cite{poursabzi2018manipulating}?
        \item \textit{Dynamic Model Updates}: Will the explanation be used to garner feedback from the stakeholder as to how the model ought to be updated to better align with their intuition? That is, how does the stakeholder interact with the model after viewing the explanation? \citet{ross2017right} attempt to develop a technique for dynamic explanations, wherein the human can guide the model towards learning the correct explanation.
    \end{itemize}
\end{enumerate}
Once desiderata are clarified, stakeholders should be consulted again.

\subsection{Concerns of Explainability}
\label{sec:trends}
While there are positive reasons to encourage explainability of ML models, we note some concerns  raised in our interviews. 

\subsubsection{On Causality}
One chief scientist told us that ``Figuring out causal factors is the holy grail of explainability.'' However, causal explanations are largely lacking in the literature, with a few exceptions \cite{causal}. 
Though non-causal explanations can still provide valid and useful interpretations of how the model works \cite{miller2018explanation}, many organizations said that they would be keen to use causal explanations if they were available. 


\subsubsection{On Privacy}
Three organizations mentioned data privacy in the context of explainability, since in some cases explanations can be used to learn about the model \cite{milli2019model,tramer2016stealing} or the training data \cite{shokri2019privacy}. 
Methods to counter these concerns have been developed. For example, \citet{harder2019interpretable} develop a methodology for training a differentially private model that generates local and global explanations using locally linear maps.

\subsubsection{On Improving Performance}
One purpose of explanations is to improve ML engineers' understanding of their models, in order to help them refine and improve performance. Since machine learning models are ``dual use'' \cite{brundage2018malicious}, we should be aware that in some settings, explanations or other tools could enable malicious users to increase capabilities and performance of undesirable systems.
For example, several organizations we talked with use explanation methods to improve their natural language processing and image recognition models for content moderation in ways that may concern some stakeholders. 

\subsubsection{Beyond Deep Learning}
Though deep learning has gained popularity in recent years, many organizations still use classical ML techniques (e.g., logistic regression, support vector machines), likely due to a need for simpler, more interpretable models \cite{rudin2019stop}. 

Many in the explainability community have focused on interpreting black-box deep learning models, even though practitioners feel that there is a dearth of model-specific techniques to understand traditional ML models. For example, one research scientist noted that, ``Many [financial institutions] use kernel-based methods on tabular data.'' As a result, there is a desire to translate explainability techniques for kernel support vector machines in genomics \cite{shrikumar2018gkmexplain} to models trained on tabular data.

Model agnostic techniques like \citet{shap} can be used for traditional models, but are ``likely overkill'' for explaining kernel-based ML models, according to one research scientist, since model-agnostic methods can be computationally expensive and lead to poorly approximated explanations.

\section{Conclusion}
\label{sec:future}
In this study, we critically examine how explanation techniques are used in practice.
We are the first, to our knowledge, to interview various organizations on how they deploy explainability in their ML workflows, concluding with salient directions for future research.
We found that while ML engineers are increasingly using explainability techniques as sanity checks during the development process, there are still significant limitations to current techniques that prevent their use to directly inform end users. 
These limitations include the need for domain experts to evaluate explanations, the risk of spurious correlations reflected in model explanations, the lack of causal intuition, and the latency in computing and showing explanations in real-time. Future research should seek to address these limitations.
We also highlighted the need for organizations to establish clear desiderata for their explanation techniques and to be cognizant of the concerns associated with explainability.
Through this analysis, we take a step towards describing explainability deployment and hope that future research builds trustworthy explainability solutions.

\section{Acknowledgments}
The authors would like to thank the following individuals for their advice, contributions, and/or support: 
Karina Alexanyan (Partnership on AI), Gagan Bansal (University of Washington), Rich Caruana (Microsoft), Amit Dhurandhar (IBM), Krishna Gade (Fiddler Labs), Konstantinos Georgatzis (QuantumBlack), Jette Henderson (CognitiveScale), Bahador Kaleghi (Element AI), Hima Lakkaraju (Harvard University), Katherine Lewis (Partnership on AI), Peter Lo (Partnership on AI), Terah Lyons (Partnership on AI), Saayeli Mukherji (Partnership on AI), Erik Pazos (QuantumBlack), Inioluwa Deborah Raji (Partnership on AI + AI Now), Nicole Rigillo (ElementAI), Francesca Rossi (IBM), Jay Turcot (Affectiva),  Kush Varshney (IBM), Dennis Wei (IBM), Edward Zhong (Baidu), Gabi Zijderveld (Affectiva), and ten other anonymous individuals.

UB acknowledges support from DeepMind via the Leverhulme Centre for the Future of Intelligence (CFI) and the Partnership on AI research fellowship.
AW acknowledges support from the David MacKay Newton research fellowship at Darwin College, The Alan Turing Institute under EPSRC grant EP/N510129/1 \& TU/B/000074, and the Leverhulme Trust via CFI.

\bibliographystyle{ACM-Reference-Format}
\bibliography{paper}


\begin{thebibliography}{71}


\ifx \showCODEN    \undefined \def \showCODEN     #1{\unskip}     \fi
\ifx \showDOI      \undefined \def \showDOI       #1{#1}\fi
\ifx \showISBNx    \undefined \def \showISBNx     #1{\unskip}     \fi
\ifx \showISBNxiii \undefined \def \showISBNxiii  #1{\unskip}     \fi
\ifx \showISSN     \undefined \def \showISSN      #1{\unskip}     \fi
\ifx \showLCCN     \undefined \def \showLCCN      #1{\unskip}     \fi
\ifx \shownote     \undefined \def \shownote      #1{#1}          \fi
\ifx \showarticletitle \undefined \def \showarticletitle #1{#1}   \fi
\ifx \showURL      \undefined \def \showURL       {\relax}        \fi
\providecommand\bibfield[2]{#2}
\providecommand\bibinfo[2]{#2}
\providecommand\natexlab[1]{#1}
\providecommand\showeprint[2][]{arXiv:#2}

\bibitem[\protect\citeauthoryear{??}{ibm}{2019}]%
        {ibm2019}
 \bibinfo{year}{2019}\natexlab{}.
\newblock \bibinfo{title}{IBM'S Principles for Data Trust and Transparency}.
\newblock
\newblock
\urldef\tempurl%
\url{https://www.ibm.com/blogs/policy/trust-principles/}
\showURL{%
\tempurl}


\bibitem[\protect\citeauthoryear{??}{msf}{2019}]%
        {msft2019}
 \bibinfo{year}{2019}\natexlab{}.
\newblock \bibinfo{title}{Our approach: Microsoft AI principles}.
\newblock
\newblock
\urldef\tempurl%
\url{https://www.microsoft.com/en-us/ai/our-approach-to-ai}
\showURL{%
\tempurl}


\bibitem[\protect\citeauthoryear{Adel, Ghahramani, and Weller}{Adel
  et~al\mbox{.}}{2018}]%
        {adel2018discovering}
\bibfield{author}{\bibinfo{person}{Tameem Adel}, \bibinfo{person}{Zoubin
  Ghahramani}, {and} \bibinfo{person}{Adrian Weller}.}
  \bibinfo{year}{2018}\natexlab{}.
\newblock \showarticletitle{Discovering interpretable representations for both
  deep generative and discriminative models}. In
  \bibinfo{booktitle}{\emph{International Conference on Machine Learning}}.
  \bibinfo{pages}{50--59}.
\newblock


\bibitem[\protect\citeauthoryear{Adel~Bargal, Zunino, Kim, Zhang, Murino, and
  Sclaroff}{Adel~Bargal et~al\mbox{.}}{2018}]%
        {adel2018excitation}
\bibfield{author}{\bibinfo{person}{Sarah Adel~Bargal}, \bibinfo{person}{Andrea
  Zunino}, \bibinfo{person}{Donghyun Kim}, \bibinfo{person}{Jianming Zhang},
  \bibinfo{person}{Vittorio Murino}, {and} \bibinfo{person}{Stan Sclaroff}.}
  \bibinfo{year}{2018}\natexlab{}.
\newblock \showarticletitle{Excitation backprop for RNNs}. In
  \bibinfo{booktitle}{\emph{Proceedings of the IEEE Conference on Computer
  Vision and Pattern Recognition}}. \bibinfo{pages}{1440--1449}.
\newblock


\bibitem[\protect\citeauthoryear{Alvarado and Waern}{Alvarado and
  Waern}{2018}]%
        {alvarado2018towards}
\bibfield{author}{\bibinfo{person}{Oscar Alvarado} {and}
  \bibinfo{person}{Annika Waern}.} \bibinfo{year}{2018}\natexlab{}.
\newblock \showarticletitle{Towards algorithmic experience: Initial efforts for
  social media contexts}. In \bibinfo{booktitle}{\emph{Proceedings of the 2018
  CHI Conference on Human Factors in Computing Systems}}. ACM,
  \bibinfo{pages}{286}.
\newblock


\bibitem[\protect\citeauthoryear{Amodei, Olah, Steinhardt, Christiano,
  Schulman, and Man{\'e}}{Amodei et~al\mbox{.}}{2016}]%
        {amodei2016concrete}
\bibfield{author}{\bibinfo{person}{Dario Amodei}, \bibinfo{person}{Chris Olah},
  \bibinfo{person}{Jacob Steinhardt}, \bibinfo{person}{Paul Christiano},
  \bibinfo{person}{John Schulman}, {and} \bibinfo{person}{Dan Man{\'e}}.}
  \bibinfo{year}{2016}\natexlab{}.
\newblock \showarticletitle{Concrete problems in AI safety}.
\newblock \bibinfo{journal}{\emph{arXiv preprint arXiv:1606.06565}}
  (\bibinfo{year}{2016}).
\newblock


\bibitem[\protect\citeauthoryear{Ancona, Ceolini, Oztireli, and Gross}{Ancona
  et~al\mbox{.}}{2018}]%
        {ancona2018towards}
\bibfield{author}{\bibinfo{person}{Marco Ancona}, \bibinfo{person}{Enea
  Ceolini}, \bibinfo{person}{Cengiz Oztireli}, {and} \bibinfo{person}{Markus
  Gross}.} \bibinfo{year}{2018}\natexlab{}.
\newblock \showarticletitle{Towards better understanding of gradient-based
  attribution methods for Deep Neural Networks}. In
  \bibinfo{booktitle}{\emph{6th {I}nternational {C}onference on {L}earning
  {R}epresentations ({ICLR} 2018)}}.
\newblock


\bibitem[\protect\citeauthoryear{Ancona, Oztireli, and Gross}{Ancona
  et~al\mbox{.}}{2019}]%
        {dasp}
\bibfield{author}{\bibinfo{person}{Marco Ancona}, \bibinfo{person}{Cengiz
  Oztireli}, {and} \bibinfo{person}{Markus Gross}.}
  \bibinfo{year}{2019}\natexlab{}.
\newblock \showarticletitle{Explaining Deep Neural Networks with a Polynomial
  Time Algorithm for Shapley Value Approximation}. In
  \bibinfo{booktitle}{\emph{Proceedings of the 36th International Conference on
  Machine Learning}} \emph{(\bibinfo{series}{Proceedings of Machine Learning
  Research})}, \bibfield{editor}{\bibinfo{person}{Kamalika Chaudhuri} {and}
  \bibinfo{person}{Ruslan Salakhutdinov}} (Eds.), Vol.~\bibinfo{volume}{97}.
  \bibinfo{publisher}{PMLR}, \bibinfo{address}{Long Beach, California, USA},
  \bibinfo{pages}{272--281}.
\newblock


\bibitem[\protect\citeauthoryear{Baehrens, Schroeter, Harmeling, Kawanabe,
  Hansen, and M{\~A}{\v{z}}ller}{Baehrens et~al\mbox{.}}{2010}]%
        {baehrens2010explain}
\bibfield{author}{\bibinfo{person}{David Baehrens}, \bibinfo{person}{Timon
  Schroeter}, \bibinfo{person}{Stefan Harmeling}, \bibinfo{person}{Motoaki
  Kawanabe}, \bibinfo{person}{Katja Hansen}, {and}
  \bibinfo{person}{Klaus-Robert M{\~A}{\v{z}}ller}.}
  \bibinfo{year}{2010}\natexlab{}.
\newblock \showarticletitle{How to explain individual classification
  decisions}.
\newblock \bibinfo{journal}{\emph{Journal of Machine Learning Research}}
  \bibinfo{volume}{11}, \bibinfo{number}{Jun} (\bibinfo{year}{2010}),
  \bibinfo{pages}{1803--1831}.
\newblock


\bibitem[\protect\citeauthoryear{Been~Kim and Koyejo}{Been~Kim and
  Koyejo}{2016}]%
        {kim2016MMD}
\bibfield{author}{\bibinfo{person}{Rajiv~Khanna Been~Kim} {and}
  \bibinfo{person}{Sanmi Koyejo}.} \bibinfo{year}{2016}\natexlab{}.
\newblock \showarticletitle{Examples are not Enough, Learn to Criticize!
  Criticism for Interpretability}. In \bibinfo{booktitle}{\emph{Advances in
  Neural Information Processing Systems}}.
\newblock


\bibitem[\protect\citeauthoryear{Bhatt, Ravikumar, and Moura}{Bhatt
  et~al\mbox{.}}{2019}]%
        {bhatt2019towards}
\bibfield{author}{\bibinfo{person}{Umang Bhatt}, \bibinfo{person}{Pradeep
  Ravikumar}, {and} \bibinfo{person}{Jos{\'e}~MF Moura}.}
  \bibinfo{year}{2019}\natexlab{}.
\newblock \showarticletitle{Towards aggregating weighted feature attributions}.
\newblock \bibinfo{journal}{\emph{arXiv preprint arXiv:1901.10040}}
  (\bibinfo{year}{2019}).
\newblock


\bibitem[\protect\citeauthoryear{Brundage, Avin, Clark, Toner, Eckersley,
  Garfinkel, Dafoe, Scharre, Zeitzoff, Filar, et~al\mbox{.}}{Brundage
  et~al\mbox{.}}{2018}]%
        {brundage2018malicious}
\bibfield{author}{\bibinfo{person}{Miles Brundage}, \bibinfo{person}{Shahar
  Avin}, \bibinfo{person}{Jack Clark}, \bibinfo{person}{Helen Toner},
  \bibinfo{person}{Peter Eckersley}, \bibinfo{person}{Ben Garfinkel},
  \bibinfo{person}{Allan Dafoe}, \bibinfo{person}{Paul Scharre},
  \bibinfo{person}{Thomas Zeitzoff}, \bibinfo{person}{Bobby Filar},
  {et~al\mbox{.}}} \bibinfo{year}{2018}\natexlab{}.
\newblock \showarticletitle{The malicious use of artificial intelligence:
  Forecasting, prevention, and mitigation}.
\newblock \bibinfo{journal}{\emph{arXiv preprint arXiv:1802.07228}}
  (\bibinfo{year}{2018}).
\newblock


\bibitem[\protect\citeauthoryear{Chattopadhyay, Manupriya, Sarkar, and
  Balasubramanian}{Chattopadhyay et~al\mbox{.}}{2019}]%
        {causal}
\bibfield{author}{\bibinfo{person}{Aditya Chattopadhyay},
  \bibinfo{person}{Piyushi Manupriya}, \bibinfo{person}{Anirban Sarkar}, {and}
  \bibinfo{person}{Vineeth~N Balasubramanian}.}
  \bibinfo{year}{2019}\natexlab{}.
\newblock \showarticletitle{Neural Network Attributions: A Causal Perspective}.
  In \bibinfo{booktitle}{\emph{Proceedings of the 36th International Conference
  on Machine Learning}} \emph{(\bibinfo{series}{Proceedings of Machine Learning
  Research})}, \bibfield{editor}{\bibinfo{person}{Kamalika Chaudhuri} {and}
  \bibinfo{person}{Ruslan Salakhutdinov}} (Eds.), Vol.~\bibinfo{volume}{97}.
  \bibinfo{publisher}{PMLR}, \bibinfo{address}{Long Beach, California, USA},
  \bibinfo{pages}{981--990}.
\newblock


\bibitem[\protect\citeauthoryear{Chen, Song, Wainwright, and Jordan}{Chen
  et~al\mbox{.}}{[n. d.]}]%
        {chen2018shapley}
\bibfield{author}{\bibinfo{person}{Jianbo Chen}, \bibinfo{person}{Le Song},
  \bibinfo{person}{Martin~J Wainwright}, {and} \bibinfo{person}{Michael~I
  Jordan}.} \bibinfo{year}{[n. d.]}\natexlab{}.
\newblock \showarticletitle{L-shapley and c-shapley: Efficient model
  interpretation for structured data}.
\newblock \bibinfo{journal}{\emph{7th {I}nternational {C}onference on
  {L}earning {R}epresentations ({ICLR} 2019)}} (\bibinfo{year}{[n. d.]}).
\newblock


\bibitem[\protect\citeauthoryear{Cook}{Cook}{1977}]%
        {cook1977detection}
\bibfield{author}{\bibinfo{person}{R~Dennis Cook}.}
  \bibinfo{year}{1977}\natexlab{}.
\newblock \showarticletitle{Detection of influential observation in linear
  regression}.
\newblock \bibinfo{journal}{\emph{Technometrics}} \bibinfo{volume}{19},
  \bibinfo{number}{1} (\bibinfo{year}{1977}), \bibinfo{pages}{15--18}.
\newblock


\bibitem[\protect\citeauthoryear{De~Fauw, Ledsam, Romera-Paredes, Nikolov,
  Tomasev, Blackwell, Askham, Glorot, O'Donoghue, Visentin,
  et~al\mbox{.}}{De~Fauw et~al\mbox{.}}{2018}]%
        {de2018clinically}
\bibfield{author}{\bibinfo{person}{Jeffrey De~Fauw}, \bibinfo{person}{Joseph~R
  Ledsam}, \bibinfo{person}{Bernardino Romera-Paredes},
  \bibinfo{person}{Stanislav Nikolov}, \bibinfo{person}{Nenad Tomasev},
  \bibinfo{person}{Sam Blackwell}, \bibinfo{person}{Harry Askham},
  \bibinfo{person}{Xavier Glorot}, \bibinfo{person}{Brendan O'Donoghue},
  \bibinfo{person}{Daniel Visentin}, {et~al\mbox{.}}}
  \bibinfo{year}{2018}\natexlab{}.
\newblock \showarticletitle{Clinically applicable deep learning for diagnosis
  and referral in retinal disease}.
\newblock \bibinfo{journal}{\emph{Nature medicine}} \bibinfo{volume}{24},
  \bibinfo{number}{9} (\bibinfo{year}{2018}), \bibinfo{pages}{1342}.
\newblock


\bibitem[\protect\citeauthoryear{Dhurandhar, Shanmugam, Luss, and
  Olsen}{Dhurandhar et~al\mbox{.}}{2018}]%
        {dhurandhar2018improving}
\bibfield{author}{\bibinfo{person}{Amit Dhurandhar},
  \bibinfo{person}{Karthikeyan Shanmugam}, \bibinfo{person}{Ronny Luss}, {and}
  \bibinfo{person}{Peder~A Olsen}.} \bibinfo{year}{2018}\natexlab{}.
\newblock \showarticletitle{Improving simple models with confidence profiles}.
  In \bibinfo{booktitle}{\emph{Advances in Neural Information Processing
  Systems}}. \bibinfo{pages}{10296--10306}.
\newblock


\bibitem[\protect\citeauthoryear{Dombrowski, Alber, Anders, Ackermann,
  M{\"u}ller, and Kessel}{Dombrowski et~al\mbox{.}}{2019}]%
        {dombrowski2019explanations}
\bibfield{author}{\bibinfo{person}{Ann-Kathrin Dombrowski},
  \bibinfo{person}{Maximilian Alber}, \bibinfo{person}{Christopher~J Anders},
  \bibinfo{person}{Marcel Ackermann}, \bibinfo{person}{Klaus-Robert
  M{\"u}ller}, {and} \bibinfo{person}{Pan Kessel}.}
  \bibinfo{year}{2019}\natexlab{}.
\newblock \showarticletitle{Explanations can be manipulated and geometry is to
  blame}.
\newblock \bibinfo{journal}{\emph{arXiv preprint arXiv:1906.07983}}
  (\bibinfo{year}{2019}).
\newblock


\bibitem[\protect\citeauthoryear{Doshi-Velez and Kim}{Doshi-Velez and
  Kim}{2017}]%
        {doshi2017towards}
\bibfield{author}{\bibinfo{person}{Finale Doshi-Velez} {and}
  \bibinfo{person}{Been Kim}.} \bibinfo{year}{2017}\natexlab{}.
\newblock \showarticletitle{Towards A Rigorous Science of Interpretable Machine
  Learning}.
\newblock  (\bibinfo{year}{2017}).
\newblock


\bibitem[\protect\citeauthoryear{DuMouchel}{DuMouchel}{2002}]%
        {dumouchel2002data}
\bibfield{author}{\bibinfo{person}{William DuMouchel}.}
  \bibinfo{year}{2002}\natexlab{}.
\newblock \showarticletitle{Data squashing: constructing summary data sets}.
\newblock In \bibinfo{booktitle}{\emph{Handbook of Massive Data Sets}}.
  \bibinfo{publisher}{Springer}, \bibinfo{pages}{579--591}.
\newblock


\bibitem[\protect\citeauthoryear{Etmann, Lunz, Maass, and Schoenlieb}{Etmann
  et~al\mbox{.}}{2019}]%
        {adv2int}
\bibfield{author}{\bibinfo{person}{Christian Etmann},
  \bibinfo{person}{Sebastian Lunz}, \bibinfo{person}{Peter Maass}, {and}
  \bibinfo{person}{Carola Schoenlieb}.} \bibinfo{year}{2019}\natexlab{}.
\newblock \showarticletitle{On the Connection Between Adversarial Robustness
  and Saliency Map Interpretability}. In \bibinfo{booktitle}{\emph{Proceedings
  of the 36th International Conference on Machine Learning}}
  \emph{(\bibinfo{series}{Proceedings of Machine Learning Research})},
  \bibfield{editor}{\bibinfo{person}{Kamalika Chaudhuri} {and}
  \bibinfo{person}{Ruslan Salakhutdinov}} (Eds.), Vol.~\bibinfo{volume}{97}.
  \bibinfo{publisher}{PMLR}, \bibinfo{address}{Long Beach, California, USA},
  \bibinfo{pages}{1823--1832}.
\newblock


\bibitem[\protect\citeauthoryear{Fong and Vedaldi}{Fong and Vedaldi}{2017}]%
        {meaningful_pert}
\bibfield{author}{\bibinfo{person}{Ruth Fong} {and} \bibinfo{person}{Andrea
  Vedaldi}.} \bibinfo{year}{2017}\natexlab{}.
\newblock \showarticletitle{Interpretable Explanations of Black Boxes by
  Meaningful Perturbation}.
\newblock \bibinfo{howpublished}{Proceedings of the 2017 IEEE International
  Conference on Computer Vision (ICCV)}.
\newblock  (\bibinfo{year}{2017}).
\newblock
\urldef\tempurl%
\url{https://doi.org/10.1109/ICCV.2017.371}
\showDOI{\tempurl}
\showeprint{arXiv:1704.03296}


\bibitem[\protect\citeauthoryear{Ghorbani, Abid, and Zou}{Ghorbani
  et~al\mbox{.}}{2019}]%
        {ghorbani2017interpretation}
\bibfield{author}{\bibinfo{person}{Amirata Ghorbani}, \bibinfo{person}{Abubakar
  Abid}, {and} \bibinfo{person}{James Zou}.} \bibinfo{year}{2019}\natexlab{}.
\newblock \showarticletitle{Interpretation of neural networks is fragile}.
\newblock \bibinfo{journal}{\emph{AAAI}} (\bibinfo{year}{2019}).
\newblock


\bibitem[\protect\citeauthoryear{Gilpin, Bau, Yuan, Bajwa, Specter, and
  Kagal}{Gilpin et~al\mbox{.}}{2018}]%
        {gilpin2018explaining}
\bibfield{author}{\bibinfo{person}{Leilani~H Gilpin}, \bibinfo{person}{David
  Bau}, \bibinfo{person}{Ben~Z Yuan}, \bibinfo{person}{Ayesha Bajwa},
  \bibinfo{person}{Michael Specter}, {and} \bibinfo{person}{Lalana Kagal}.}
  \bibinfo{year}{2018}\natexlab{}.
\newblock \showarticletitle{Explaining explanations: An overview of
  interpretability of machine learning}. In \bibinfo{booktitle}{\emph{2018 IEEE
  5th International Conference on data science and advanced analytics (DSAA)}}.
  IEEE, \bibinfo{pages}{80--89}.
\newblock


\bibitem[\protect\citeauthoryear{Harder, Bauer, and Park}{Harder
  et~al\mbox{.}}{2019}]%
        {harder2019interpretable}
\bibfield{author}{\bibinfo{person}{Frederik Harder}, \bibinfo{person}{Matthias
  Bauer}, {and} \bibinfo{person}{Mijung Park}.}
  \bibinfo{year}{2019}\natexlab{}.
\newblock \showarticletitle{Interpretable and Differentially Private
  Predictions}.
\newblock \bibinfo{journal}{\emph{arXiv preprint arXiv:1906.02004}}
  (\bibinfo{year}{2019}).
\newblock


\bibitem[\protect\citeauthoryear{Heaton, Polson, and Witte}{Heaton
  et~al\mbox{.}}{2016}]%
        {heaton2016deep}
\bibfield{author}{\bibinfo{person}{JB Heaton}, \bibinfo{person}{Nicholas~G
  Polson}, {and} \bibinfo{person}{Jan~Hendrik Witte}.}
  \bibinfo{year}{2016}\natexlab{}.
\newblock \showarticletitle{Deep learning in finance}.
\newblock \bibinfo{journal}{\emph{arXiv preprint arXiv:1602.06561}}
  (\bibinfo{year}{2016}).
\newblock


\bibitem[\protect\citeauthoryear{Holland}{Holland}{1986}]%
        {rubin}
\bibfield{author}{\bibinfo{person}{Paul~W. Holland}.}
  \bibinfo{year}{1986}\natexlab{}.
\newblock \showarticletitle{Statistics and Causal Inference}.
\newblock \bibinfo{journal}{\emph{J. Amer. Statist. Assoc.}}
  \bibinfo{volume}{81}, \bibinfo{number}{396} (\bibinfo{year}{1986}),
  \bibinfo{pages}{945--960}.
\newblock


\bibitem[\protect\citeauthoryear{Holstein, Wortman~Vaughan, Daum{\'e}~III,
  Dudik, and Wallach}{Holstein et~al\mbox{.}}{2019}]%
        {holstein2019improving}
\bibfield{author}{\bibinfo{person}{Kenneth Holstein}, \bibinfo{person}{Jennifer
  Wortman~Vaughan}, \bibinfo{person}{Hal Daum{\'e}~III}, \bibinfo{person}{Miro
  Dudik}, {and} \bibinfo{person}{Hanna Wallach}.}
  \bibinfo{year}{2019}\natexlab{}.
\newblock \showarticletitle{Improving fairness in machine learning systems:
  What do industry practitioners need?}. In
  \bibinfo{booktitle}{\emph{Proceedings of the 2019 CHI Conference on Human
  Factors in Computing Systems}}. ACM, \bibinfo{pages}{600}.
\newblock


\bibitem[\protect\citeauthoryear{Hooker and Mentch}{Hooker and Mentch}{2019}]%
        {hooker2019please}
\bibfield{author}{\bibinfo{person}{Giles Hooker} {and} \bibinfo{person}{Lucas
  Mentch}.} \bibinfo{year}{2019}\natexlab{}.
\newblock \showarticletitle{Please Stop Permuting Features: An Explanation and
  Alternatives}.
\newblock \bibinfo{journal}{\emph{arXiv preprint arXiv:1905.03151}}
  (\bibinfo{year}{2019}).
\newblock


\bibitem[\protect\citeauthoryear{Ilyas, Santurkar, Tsipras, Engstrom, Tran, and
  Madry}{Ilyas et~al\mbox{.}}{2019}]%
        {ilyas2019adversarial}
\bibfield{author}{\bibinfo{person}{Andrew Ilyas}, \bibinfo{person}{Shibani
  Santurkar}, \bibinfo{person}{Dimitris Tsipras}, \bibinfo{person}{Logan
  Engstrom}, \bibinfo{person}{Brandon Tran}, {and} \bibinfo{person}{Aleksander
  Madry}.} \bibinfo{year}{2019}\natexlab{}.
\newblock \bibinfo{title}{Adversarial Examples Are Not Bugs, They Are
  Features}.
\newblock
\newblock
\urldef\tempurl%
\url{http://arxiv.org/abs/1905.02175}
\showURL{%
\tempurl}
\newblock
\shownote{cite arxiv:1905.02175.}


\bibitem[\protect\citeauthoryear{Kim, Wattenberg, Gilmer, Cai, Wexler, Viegas,
  and Sayres}{Kim et~al\mbox{.}}{2017}]%
        {kim2017interpretability}
\bibfield{author}{\bibinfo{person}{Been Kim}, \bibinfo{person}{Martin
  Wattenberg}, \bibinfo{person}{Justin Gilmer}, \bibinfo{person}{Carrie Cai},
  \bibinfo{person}{James Wexler}, \bibinfo{person}{Fernanda Viegas}, {and}
  \bibinfo{person}{Rory Sayres}.} \bibinfo{year}{2017}\natexlab{}.
\newblock \showarticletitle{Interpretability beyond feature attribution:
  Quantitative testing with concept activation vectors (tcav)}.
\newblock \bibinfo{journal}{\emph{arXiv preprint arXiv:1711.11279}}
  (\bibinfo{year}{2017}).
\newblock


\bibitem[\protect\citeauthoryear{Koh and Liang}{Koh and Liang}{2017}]%
        {koh2017understanding}
\bibfield{author}{\bibinfo{person}{Pang~Wei Koh} {and} \bibinfo{person}{Percy
  Liang}.} \bibinfo{year}{2017}\natexlab{}.
\newblock \showarticletitle{Understanding black-box predictions via influence
  functions}. In \bibinfo{booktitle}{\emph{Proceedings of the 34th
  International Conference on Machine Learning-Volume 70 ({ICML} 2017)}}.
  Journal of Machine Learning Research, \bibinfo{pages}{1885--1894}.
\newblock


\bibitem[\protect\citeauthoryear{Lepri, Oliver, Letouz{\'e}, Pentland, and
  Vinck}{Lepri et~al\mbox{.}}{2018}]%
        {lepri2018fair}
\bibfield{author}{\bibinfo{person}{Bruno Lepri}, \bibinfo{person}{Nuria
  Oliver}, \bibinfo{person}{Emmanuel Letouz{\'e}}, \bibinfo{person}{Alex
  Pentland}, {and} \bibinfo{person}{Patrick Vinck}.}
  \bibinfo{year}{2018}\natexlab{}.
\newblock \showarticletitle{Fair, transparent, and accountable algorithmic
  decision-making processes}.
\newblock \bibinfo{journal}{\emph{Philosophy \& Technology}}
  \bibinfo{volume}{31}, \bibinfo{number}{4} (\bibinfo{year}{2018}),
  \bibinfo{pages}{611--627}.
\newblock


\bibitem[\protect\citeauthoryear{Lundberg and Lee}{Lundberg and Lee}{2017}]%
        {shap}
\bibfield{author}{\bibinfo{person}{Scott~M Lundberg} {and}
  \bibinfo{person}{Su-In Lee}.} \bibinfo{year}{2017}\natexlab{}.
\newblock \showarticletitle{A Unified Approach to Interpreting Model
  Predictions}.
\newblock In \bibinfo{booktitle}{\emph{Advances in Neural Information
  Processing Systems 30 ({NeurIPS} 2017)}},
  \bibfield{editor}{\bibinfo{person}{I.~Guyon}, \bibinfo{person}{U.~V.
  Luxburg}, \bibinfo{person}{S.~Bengio}, \bibinfo{person}{H.~Wallach},
  \bibinfo{person}{R.~Fergus}, \bibinfo{person}{S.~Vishwanathan}, {and}
  \bibinfo{person}{R.~Garnett}} (Eds.). \bibinfo{publisher}{Curran Associates,
  Inc.}, \bibinfo{pages}{4765--4774}.
\newblock


\bibitem[\protect\citeauthoryear{Lundberg, Nair, Vavilala, Horibe, Eisses,
  Adams, Liston, Low, Newman, Kim, et~al\mbox{.}}{Lundberg
  et~al\mbox{.}}{2018}]%
        {lundberg2018explainable}
\bibfield{author}{\bibinfo{person}{Scott~M Lundberg}, \bibinfo{person}{Bala
  Nair}, \bibinfo{person}{Monica~S Vavilala}, \bibinfo{person}{Mayumi Horibe},
  \bibinfo{person}{Michael~J Eisses}, \bibinfo{person}{Trevor Adams},
  \bibinfo{person}{David~E Liston}, \bibinfo{person}{Daniel King-Wai Low},
  \bibinfo{person}{Shu-Fang Newman}, \bibinfo{person}{Jerry Kim},
  {et~al\mbox{.}}} \bibinfo{year}{2018}\natexlab{}.
\newblock \showarticletitle{Explainable machine-learning predictions for the
  prevention of hypoxaemia during surgery}.
\newblock \bibinfo{journal}{\emph{Nature biomedical engineering}}
  \bibinfo{volume}{2}, \bibinfo{number}{10} (\bibinfo{year}{2018}),
  \bibinfo{pages}{749}.
\newblock


\bibitem[\protect\citeauthoryear{Madry, Makelov, Schmidt, Tsipras, and
  Vladu}{Madry et~al\mbox{.}}{2017}]%
        {madry2017towards}
\bibfield{author}{\bibinfo{person}{Aleksander Madry},
  \bibinfo{person}{Aleksandar Makelov}, \bibinfo{person}{Ludwig Schmidt},
  \bibinfo{person}{Dimitris Tsipras}, {and} \bibinfo{person}{Adrian Vladu}.}
  \bibinfo{year}{2017}\natexlab{}.
\newblock \showarticletitle{Towards deep learning models resistant to
  adversarial attacks}.
\newblock \bibinfo{journal}{\emph{arXiv preprint arXiv:1706.06083}}
  (\bibinfo{year}{2017}).
\newblock


\bibitem[\protect\citeauthoryear{Miller}{Miller}{2018}]%
        {miller2018explanation}
\bibfield{author}{\bibinfo{person}{Tim Miller}.}
  \bibinfo{year}{2018}\natexlab{}.
\newblock \showarticletitle{Explanation in artificial intelligence: Insights
  from the social sciences}.
\newblock \bibinfo{journal}{\emph{Artificial Intelligence}}
  (\bibinfo{year}{2018}).
\newblock


\bibitem[\protect\citeauthoryear{Milli, Schmidt, Dragan, and Hardt}{Milli
  et~al\mbox{.}}{2019}]%
        {milli2019model}
\bibfield{author}{\bibinfo{person}{Smitha Milli}, \bibinfo{person}{Ludwig
  Schmidt}, \bibinfo{person}{Anca Dragan}, {and} \bibinfo{person}{Moritz
  Hardt}.} \bibinfo{year}{2019}\natexlab{}.
\newblock \showarticletitle{Model Reconstruction from Model Explanations}.
\newblock \bibinfo{journal}{\emph{In Proceedings of ACM FAT* 2019}}
  (\bibinfo{year}{2019}).
\newblock


\bibitem[\protect\citeauthoryear{Mitchell, Wu, Zaldivar, Barnes, Vasserman,
  Hutchinson, Spitzer, Raji, and Gebru}{Mitchell et~al\mbox{.}}{2019}]%
        {mitchell2019model}
\bibfield{author}{\bibinfo{person}{Margaret Mitchell}, \bibinfo{person}{Simone
  Wu}, \bibinfo{person}{Andrew Zaldivar}, \bibinfo{person}{Parker Barnes},
  \bibinfo{person}{Lucy Vasserman}, \bibinfo{person}{Ben Hutchinson},
  \bibinfo{person}{Elena Spitzer}, \bibinfo{person}{Inioluwa~Deborah Raji},
  {and} \bibinfo{person}{Timnit Gebru}.} \bibinfo{year}{2019}\natexlab{}.
\newblock \showarticletitle{Model cards for model reporting}. In
  \bibinfo{booktitle}{\emph{Proceedings of the Conference on Fairness,
  Accountability, and Transparency}}. ACM, \bibinfo{pages}{220--229}.
\newblock


\bibitem[\protect\citeauthoryear{Mittelstadt, Russell, and Wachter}{Mittelstadt
  et~al\mbox{.}}{2019}]%
        {mittelstadt2019explaining}
\bibfield{author}{\bibinfo{person}{Brent Mittelstadt}, \bibinfo{person}{Chris
  Russell}, {and} \bibinfo{person}{Sandra Wachter}.}
  \bibinfo{year}{2019}\natexlab{}.
\newblock \showarticletitle{Explaining explanations in AI}. In
  \bibinfo{booktitle}{\emph{Proceedings of the conference on fairness,
  accountability, and transparency}}. ACM, \bibinfo{pages}{279--288}.
\newblock


\bibitem[\protect\citeauthoryear{Montavon, Lapuschkin, Binder, Samek, and
  M{\"u}ller}{Montavon et~al\mbox{.}}{2017}]%
        {dtd}
\bibfield{author}{\bibinfo{person}{Gr{\'e}goire Montavon},
  \bibinfo{person}{Sebastian Lapuschkin}, \bibinfo{person}{Alexander Binder},
  \bibinfo{person}{Wojciech Samek}, {and} \bibinfo{person}{Klaus-Robert
  M{\"u}ller}.} \bibinfo{year}{2017}\natexlab{}.
\newblock \showarticletitle{Explaining nonlinear classification decisions with
  deep taylor decomposition}.
\newblock \bibinfo{journal}{\emph{Pattern Recognition}}  \bibinfo{volume}{65}
  (\bibinfo{year}{2017}), \bibinfo{pages}{211--222}.
\newblock


\bibitem[\protect\citeauthoryear{Niu, Qiao, Li, and Huang}{Niu
  et~al\mbox{.}}{2018}]%
        {niu2018word}
\bibfield{author}{\bibinfo{person}{Yilin Niu}, \bibinfo{person}{Chao Qiao},
  \bibinfo{person}{Hang Li}, {and} \bibinfo{person}{Minlie Huang}.}
  \bibinfo{year}{2018}\natexlab{}.
\newblock \showarticletitle{Word Embedding based Edit Distance}.
\newblock \bibinfo{journal}{\emph{arXiv preprint arXiv:1810.10752}}
  (\bibinfo{year}{2018}).
\newblock


\bibitem[\protect\citeauthoryear{of~Governors of~the Federal
  Reserve~System}{of~Governors of~the Federal Reserve~System}{2011}]%
        {sr11-7}
\bibfield{author}{\bibinfo{person}{Board of~Governors of~the Federal
  Reserve~System}.} \bibinfo{year}{2011}\natexlab{}.
\newblock \showarticletitle{Supervisory Guidance on Model Risk Management}.
\newblock
  \bibinfo{journal}{\emph{https://www.federalreserve.gov/supervisionreg/srletters/sr1107a1.pdf}}
  (\bibinfo{year}{2011}).
\newblock


\bibitem[\protect\citeauthoryear{O'Neill}{O'Neill}{2018}]%
        {o2018linking}
\bibfield{author}{\bibinfo{person}{Onora O'Neill}.}
  \bibinfo{year}{2018}\natexlab{}.
\newblock \showarticletitle{Linking trust to trustworthiness}.
\newblock \bibinfo{journal}{\emph{International Journal of Philosophical
  Studies}} \bibinfo{volume}{26}, \bibinfo{number}{2} (\bibinfo{year}{2018}),
  \bibinfo{pages}{293--300}.
\newblock


\bibitem[\protect\citeauthoryear{Parliament and of~European~Union}{Parliament
  and of~European~Union}{2018}]%
        {gdpr}
\bibfield{author}{\bibinfo{person}{European Parliament} {and}
  \bibinfo{person}{Council of European~Union}.}
  \bibinfo{year}{2018}\natexlab{}.
\newblock \showarticletitle{European Union General Data Protection Regulation,
  Articles 13-15}.
\newblock \bibinfo{journal}{\emph{http://www.privacy-regulation.eu/en/13.htm}}
  (\bibinfo{year}{2018}).
\newblock


\bibitem[\protect\citeauthoryear{Pearl}{Pearl}{2000}]%
        {pearl2000causality}
\bibfield{author}{\bibinfo{person}{Judea Pearl}.}
  \bibinfo{year}{2000}\natexlab{}.
\newblock \bibinfo{booktitle}{\emph{Causality: models, reasoning and
  inference}}. Vol.~\bibinfo{volume}{29}.
\newblock \bibinfo{publisher}{Springer}.
\newblock


\bibitem[\protect\citeauthoryear{Pinto, Sampaio, and Bizarro}{Pinto
  et~al\mbox{.}}{2019}]%
        {pinto2019automatic}
\bibfield{author}{\bibinfo{person}{F{\'a}bio Pinto}, \bibinfo{person}{Marco~OP
  Sampaio}, {and} \bibinfo{person}{Pedro Bizarro}.}
  \bibinfo{year}{2019}\natexlab{}.
\newblock \showarticletitle{Automatic Model Monitoring for Data Streams}.
\newblock \bibinfo{journal}{\emph{arXiv preprint arXiv:1908.04240}}
  (\bibinfo{year}{2019}).
\newblock


\bibitem[\protect\citeauthoryear{Poursabzi-Sangdeh, Goldstein, Hofman, Vaughan,
  and Wallach}{Poursabzi-Sangdeh et~al\mbox{.}}{2018}]%
        {poursabzi2018manipulating}
\bibfield{author}{\bibinfo{person}{Forough Poursabzi-Sangdeh},
  \bibinfo{person}{Daniel~G Goldstein}, \bibinfo{person}{Jake~M Hofman},
  \bibinfo{person}{Jennifer~Wortman Vaughan}, {and} \bibinfo{person}{Hanna
  Wallach}.} \bibinfo{year}{2018}\natexlab{}.
\newblock \showarticletitle{Manipulating and measuring model interpretability}.
\newblock \bibinfo{journal}{\emph{arXiv preprint arXiv:1802.07810}}
  (\bibinfo{year}{2018}).
\newblock


\bibitem[\protect\citeauthoryear{Preece, Harborne, Braines, Tomsett, and
  Chakraborty}{Preece et~al\mbox{.}}{2018}]%
        {preece2018stakeholders}
\bibfield{author}{\bibinfo{person}{Alun Preece}, \bibinfo{person}{Dan
  Harborne}, \bibinfo{person}{Dave Braines}, \bibinfo{person}{Richard Tomsett},
  {and} \bibinfo{person}{Supriyo Chakraborty}.}
  \bibinfo{year}{2018}\natexlab{}.
\newblock \showarticletitle{Stakeholders in explainable AI}.
\newblock \bibinfo{journal}{\emph{arXiv preprint arXiv:1810.00184}}
  (\bibinfo{year}{2018}).
\newblock


\bibitem[\protect\citeauthoryear{Ribeiro, Singh, and Guestrin}{Ribeiro
  et~al\mbox{.}}{2016}]%
        {ribeiro2016should}
\bibfield{author}{\bibinfo{person}{Marco~Tulio Ribeiro},
  \bibinfo{person}{Sameer Singh}, {and} \bibinfo{person}{Carlos Guestrin}.}
  \bibinfo{year}{2016}\natexlab{}.
\newblock \showarticletitle{Why should i trust you?: Explaining the predictions
  of any classifier}. In \bibinfo{booktitle}{\emph{Proceedings of the 22nd
  {ACM} {SIGKDD} {I}nternational {C}onference on {K}nowledge {D}iscovery and
  {D}ata {M}ining}}. ACM, \bibinfo{pages}{1135--1144}.
\newblock


\bibitem[\protect\citeauthoryear{Ross, Hughes, and Doshi-Velez}{Ross
  et~al\mbox{.}}{2017}]%
        {ross2017right}
\bibfield{author}{\bibinfo{person}{Andrew~Slavin Ross},
  \bibinfo{person}{Michael~C Hughes}, {and} \bibinfo{person}{Finale
  Doshi-Velez}.} \bibinfo{year}{2017}\natexlab{}.
\newblock \showarticletitle{Right for the right reasons: training
  differentiable models by constraining their explanations}. In
  \bibinfo{booktitle}{\emph{Proceedings of the 26th International Joint
  Conference on Artificial Intelligence}}. AAAI Press,
  \bibinfo{pages}{2662--2670}.
\newblock


\bibitem[\protect\citeauthoryear{Rudin}{Rudin}{2019}]%
        {rudin2019stop}
\bibfield{author}{\bibinfo{person}{Cynthia Rudin}.}
  \bibinfo{year}{2019}\natexlab{}.
\newblock \showarticletitle{Stop explaining black box machine learning models
  for high stakes decisions and use interpretable models instead}.
\newblock \bibinfo{journal}{\emph{Nature Machine Intelligence}}
  \bibinfo{volume}{1}, \bibinfo{number}{5} (\bibinfo{year}{2019}),
  \bibinfo{pages}{206}.
\newblock


\bibitem[\protect\citeauthoryear{Selbst and Barocas}{Selbst and
  Barocas}{2018}]%
        {selbst2018intuitive}
\bibfield{author}{\bibinfo{person}{Andrew~D Selbst} {and}
  \bibinfo{person}{Solon Barocas}.} \bibinfo{year}{2018}\natexlab{}.
\newblock \showarticletitle{The intuitive appeal of explainable machines}.
\newblock \bibinfo{journal}{\emph{Fordham L. Rev.}}  \bibinfo{volume}{87}
  (\bibinfo{year}{2018}), \bibinfo{pages}{1085}.
\newblock


\bibitem[\protect\citeauthoryear{Shapley}{Shapley}{1953}]%
        {shapley52}
\bibfield{author}{\bibinfo{person}{Lloyd~S Shapley}.}
  \bibinfo{year}{1953}\natexlab{}.
\newblock \showarticletitle{A Value for n-Person Games}.
\newblock In \bibinfo{booktitle}{\emph{Contributions to the Theory of Games
  II}}. \bibinfo{pages}{307--317}.
\newblock


\bibitem[\protect\citeauthoryear{Sharma, Henderson, and Ghosh}{Sharma
  et~al\mbox{.}}{2019}]%
        {sharma2019certifai}
\bibfield{author}{\bibinfo{person}{Shubham Sharma}, \bibinfo{person}{Jette
  Henderson}, {and} \bibinfo{person}{Joydeep Ghosh}.}
  \bibinfo{year}{2019}\natexlab{}.
\newblock \showarticletitle{CERTIFAI: Counterfactual Explanations for
  Robustness, Transparency, Interpretability, and Fairness of Artificial
  Intelligence models}.
\newblock \bibinfo{journal}{\emph{arXiv preprint arXiv:1905.07857}}
  (\bibinfo{year}{2019}).
\newblock


\bibitem[\protect\citeauthoryear{Shokri, Strobel, and Zick}{Shokri
  et~al\mbox{.}}{2019}]%
        {shokri2019privacy}
\bibfield{author}{\bibinfo{person}{Reza Shokri}, \bibinfo{person}{Martin
  Strobel}, {and} \bibinfo{person}{Yair Zick}.}
  \bibinfo{year}{2019}\natexlab{}.
\newblock \showarticletitle{Privacy Risks of Explaining Machine Learning
  Models}.
\newblock \bibinfo{journal}{\emph{arXiv preprint arXiv:1907.00164}}
  (\bibinfo{year}{2019}).
\newblock


\bibitem[\protect\citeauthoryear{Shrikumar, Greenside, and Kundaje}{Shrikumar
  et~al\mbox{.}}{2017}]%
        {shrikumar2017learning}
\bibfield{author}{\bibinfo{person}{Avanti Shrikumar}, \bibinfo{person}{Peyton
  Greenside}, {and} \bibinfo{person}{Anshul Kundaje}.}
  \bibinfo{year}{2017}\natexlab{}.
\newblock \showarticletitle{Learning important features through propagating
  activation differences}. In \bibinfo{booktitle}{\emph{Proceedings of the 34th
  International Conference on Machine Learning-Volume 70 ({ICML} 2017)}}.
  Journal of Machine Learning Research, \bibinfo{pages}{3145--3153}.
\newblock


\bibitem[\protect\citeauthoryear{Shrikumar, Prakash, and Kundaje}{Shrikumar
  et~al\mbox{.}}{2018}]%
        {shrikumar2018gkmexplain}
\bibfield{author}{\bibinfo{person}{Avanti Shrikumar}, \bibinfo{person}{Eva
  Prakash}, {and} \bibinfo{person}{Anshul Kundaje}.}
  \bibinfo{year}{2018}\natexlab{}.
\newblock \showarticletitle{Gkmexplain: Fast and Accurate Interpretation of
  Nonlinear Gapped k-mer Support Vector Machines Using Integrated Gradients}.
\newblock \bibinfo{journal}{\emph{BioRxiv}} (\bibinfo{year}{2018}),
  \bibinfo{pages}{457606}.
\newblock


\bibitem[\protect\citeauthoryear{Singla, Wallace, Feng, and Feizi}{Singla
  et~al\mbox{.}}{2019}]%
        {second}
\bibfield{author}{\bibinfo{person}{Sahil Singla}, \bibinfo{person}{Eric
  Wallace}, \bibinfo{person}{Shi Feng}, {and} \bibinfo{person}{Soheil Feizi}.}
  \bibinfo{year}{2019}\natexlab{}.
\newblock \showarticletitle{Understanding Impacts of High-Order Loss
  Approximations and Features in Deep Learning Interpretation}. In
  \bibinfo{booktitle}{\emph{Proceedings of the 36th International Conference on
  Machine Learning}} \emph{(\bibinfo{series}{Proceedings of Machine Learning
  Research})}, \bibfield{editor}{\bibinfo{person}{Kamalika Chaudhuri} {and}
  \bibinfo{person}{Ruslan Salakhutdinov}} (Eds.), Vol.~\bibinfo{volume}{97}.
  \bibinfo{publisher}{PMLR}, \bibinfo{address}{Long Beach, California, USA},
  \bibinfo{pages}{5848--5856}.
\newblock


\bibitem[\protect\citeauthoryear{Smilkov, Thorat, Kim, Vi{\'e}gas, and
  Wattenberg}{Smilkov et~al\mbox{.}}{2017}]%
        {smilkov2017smoothgrad}
\bibfield{author}{\bibinfo{person}{Daniel Smilkov}, \bibinfo{person}{Nikhil
  Thorat}, \bibinfo{person}{Been Kim}, \bibinfo{person}{Fernanda Vi{\'e}gas},
  {and} \bibinfo{person}{Martin Wattenberg}.} \bibinfo{year}{2017}\natexlab{}.
\newblock \showarticletitle{Smoothgrad: removing noise by adding noise}.
\newblock \bibinfo{journal}{\emph{arXiv preprint arXiv:1706.03825}}
  (\bibinfo{year}{2017}).
\newblock


\bibitem[\protect\citeauthoryear{{\v{S}}trumbelj and Kononenko}{{\v{S}}trumbelj
  and Kononenko}{2014}]%
        {vstrumbelj2014explaining}
\bibfield{author}{\bibinfo{person}{Erik {\v{S}}trumbelj} {and}
  \bibinfo{person}{Igor Kononenko}.} \bibinfo{year}{2014}\natexlab{}.
\newblock \showarticletitle{Explaining prediction models and individual
  predictions with feature contributions}.
\newblock \bibinfo{journal}{\emph{{K}nowledge and {I}nformation {S}ystems}}
  \bibinfo{volume}{41}, \bibinfo{number}{3} (\bibinfo{year}{2014}),
  \bibinfo{pages}{647--665}.
\newblock


\bibitem[\protect\citeauthoryear{Sundararajan, Taly, and Yan}{Sundararajan
  et~al\mbox{.}}{2017}]%
        {sundararajan2017axiomatic}
\bibfield{author}{\bibinfo{person}{Mukund Sundararajan}, \bibinfo{person}{Ankur
  Taly}, {and} \bibinfo{person}{Qiqi Yan}.} \bibinfo{year}{2017}\natexlab{}.
\newblock \showarticletitle{Axiomatic attribution for deep networks}. In
  \bibinfo{booktitle}{\emph{Proceedings of the 34th International Conference on
  Machine Learning-Volume 70 ({ICML} 2017)}}. Journal of Machine Learning
  Research, \bibinfo{pages}{3319--3328}.
\newblock


\bibitem[\protect\citeauthoryear{Tram{\`e}r, Zhang, Juels, Reiter, and
  Ristenpart}{Tram{\`e}r et~al\mbox{.}}{2016}]%
        {tramer2016stealing}
\bibfield{author}{\bibinfo{person}{Florian Tram{\`e}r}, \bibinfo{person}{Fan
  Zhang}, \bibinfo{person}{Ari Juels}, \bibinfo{person}{Michael~K Reiter},
  {and} \bibinfo{person}{Thomas Ristenpart}.} \bibinfo{year}{2016}\natexlab{}.
\newblock \showarticletitle{Stealing machine learning models via prediction
  apis}. In \bibinfo{booktitle}{\emph{25th $\{$USENIX$\}$ Security Symposium
  ($\{$USENIX$\}$ Security 16)}}. \bibinfo{pages}{601--618}.
\newblock


\bibitem[\protect\citeauthoryear{Tsipras, Santurkar, Engstrom, Turner, and
  Madry}{Tsipras et~al\mbox{.}}{2019}]%
        {tsipras2018robustness}
\bibfield{author}{\bibinfo{person}{Dimitris Tsipras}, \bibinfo{person}{Shibani
  Santurkar}, \bibinfo{person}{Logan Engstrom}, \bibinfo{person}{Alexander
  Turner}, {and} \bibinfo{person}{Aleksander Madry}.}
  \bibinfo{year}{2019}\natexlab{}.
\newblock \showarticletitle{Robustness May Be at Odds with Accuracy}. In
  \bibinfo{booktitle}{\emph{International Conference on Learning
  Representations}}.
\newblock
\urldef\tempurl%
\url{https://openreview.net/forum?id=SyxAb30cY7}
\showURL{%
\tempurl}


\bibitem[\protect\citeauthoryear{Ustun, Spangher, and Liu}{Ustun
  et~al\mbox{.}}{2019}]%
        {ustun2019actionable}
\bibfield{author}{\bibinfo{person}{Berk Ustun}, \bibinfo{person}{Alexander
  Spangher}, {and} \bibinfo{person}{Yang Liu}.}
  \bibinfo{year}{2019}\natexlab{}.
\newblock \showarticletitle{Actionable recourse in linear classification}. In
  \bibinfo{booktitle}{\emph{Proceedings of the Conference on Fairness,
  Accountability, and Transparency}}. ACM, \bibinfo{pages}{10--19}.
\newblock


\bibitem[\protect\citeauthoryear{Wachter, Mittelstadt, and Russell}{Wachter
  et~al\mbox{.}}{2017}]%
        {wachter2017counterfactual}
\bibfield{author}{\bibinfo{person}{Sandra Wachter}, \bibinfo{person}{Brent
  Mittelstadt}, {and} \bibinfo{person}{Chris Russell}.}
  \bibinfo{year}{2017}\natexlab{}.
\newblock \showarticletitle{Counterfactual Explanations without Opening the
  Black Box: Automated Decisions and the GPDR}.
\newblock \bibinfo{journal}{\emph{Harv. JL \& Tech.}}  \bibinfo{volume}{31}
  (\bibinfo{year}{2017}), \bibinfo{pages}{841}.
\newblock


\bibitem[\protect\citeauthoryear{Weller}{Weller}{2019}]%
        {weller2019transparency}
\bibfield{author}{\bibinfo{person}{Adrian Weller}.}
  \bibinfo{year}{2019}\natexlab{}.
\newblock \showarticletitle{Transparency: motivations and challenges}.
\newblock In \bibinfo{booktitle}{\emph{Explainable AI: Interpreting, Explaining
  and Visualizing Deep Learning}}. \bibinfo{publisher}{Springer},
  \bibinfo{pages}{23--40}.
\newblock


\bibitem[\protect\citeauthoryear{Wexler, Pushkarna, Bolukbasi, Wattenberg,
  Viegas, and Wilson}{Wexler et~al\mbox{.}}{2019}]%
        {wexler2019if}
\bibfield{author}{\bibinfo{person}{James Wexler}, \bibinfo{person}{Mahima
  Pushkarna}, \bibinfo{person}{Tolga Bolukbasi}, \bibinfo{person}{Martin
  Wattenberg}, \bibinfo{person}{Fernanda Viegas}, {and} \bibinfo{person}{Jimbo
  Wilson}.} \bibinfo{year}{2019}\natexlab{}.
\newblock \showarticletitle{The What-If Tool: Interactive Probing of Machine
  Learning Models}.
\newblock \bibinfo{journal}{\emph{arXiv preprint arXiv:1907.04135}}
  (\bibinfo{year}{2019}).
\newblock


\bibitem[\protect\citeauthoryear{Yeh, Hsieh, Suggala, Inouye, and
  Ravikumar}{Yeh et~al\mbox{.}}{2019}]%
        {yeh2019sensitive}
\bibfield{author}{\bibinfo{person}{Chih-Kuan Yeh}, \bibinfo{person}{Cheng-Yu
  Hsieh}, \bibinfo{person}{Arun~Sai Suggala}, \bibinfo{person}{David Inouye},
  {and} \bibinfo{person}{Pradeep Ravikumar}.} \bibinfo{year}{2019}\natexlab{}.
\newblock \showarticletitle{How Sensitive are Sensitivity-Based Explanations?}
\newblock \bibinfo{journal}{\emph{arXiv preprint arXiv:1901.09392}}
  (\bibinfo{year}{2019}).
\newblock


\bibitem[\protect\citeauthoryear{Zhang, Bargal, Lin, Brandt, Shen, and
  Sclaroff}{Zhang et~al\mbox{.}}{2018}]%
        {zhang2018top}
\bibfield{author}{\bibinfo{person}{Jianming Zhang}, \bibinfo{person}{Sarah~Adel
  Bargal}, \bibinfo{person}{Zhe Lin}, \bibinfo{person}{Jonathan Brandt},
  \bibinfo{person}{Xiaohui Shen}, {and} \bibinfo{person}{Stan Sclaroff}.}
  \bibinfo{year}{2018}\natexlab{}.
\newblock \showarticletitle{Top-down neural attention by excitation backprop}.
\newblock \bibinfo{journal}{\emph{International Journal of Computer Vision}}
  \bibinfo{volume}{126}, \bibinfo{number}{10} (\bibinfo{year}{2018}),
  \bibinfo{pages}{1084--1102}.
\newblock


\bibitem[\protect\citeauthoryear{Zhang, Song, Sun, Tan, and Udell}{Zhang
  et~al\mbox{.}}{2019}]%
        {badLIME}
\bibfield{author}{\bibinfo{person}{Yujia Zhang}, \bibinfo{person}{Kuangyan
  Song}, \bibinfo{person}{Yiming Sun}, \bibinfo{person}{Sarah Tan}, {and}
  \bibinfo{person}{Madeleine Udell}.} \bibinfo{year}{2019}\natexlab{}.
\newblock \bibinfo{title}{"Why Should You Trust My Explanation?" Understanding
  Uncertainty in LIME Explanations}.
\newblock
\newblock
\showeprint{arXiv:1904.12991}


\end{thebibliography}

\end{document}